\documentclass[runningheads]{llncs}
\usepackage{graphicx}
\usepackage{booktabs}
\usepackage{array} %
\usepackage[hidelinks,colorlinks=true,allcolors=blue]{hyperref}
\usepackage{xcolor}
\usepackage{amsmath, amssymb}
\usepackage{soul}
\usepackage{multirow}
\usepackage{array}
\usepackage{float}
\usepackage[misc,geometry]{ifsym}
\usepackage{blindtext}

\usepackage{siunitx}   %
\usepackage{caption}   %

\newcolumntype{C}[1]{>{\centering\let\newline\\\arraybackslash\hspace{0pt}}m{#1}}
\usepackage{color}

\newcommand{\etal}{\textit{et al.}}

\begin{document}
\title{StylusAI: Stylistic Adaptation for Robust German Handwritten Text Generation}
\titlerunning{StylusAI}

\author{
    Nauman Riaz (\Letter) \inst{1,2} \orcidID{0009-0000-1416-3220} \and
    Saifullah Saifullah \inst{1,2} \orcidID{0000-0003-3098-2458} \and
    Stefan Agne\inst{1,3} \orcidID{0000-0002-9697-4285} \and
    Andreas Dengel\inst{1,2} \orcidID{0000-0002-6100-8255} \and
    Sheraz Ahmed\inst{1,3} \orcidID{0000-0002-4239-6520}}
\authorrunning{Nauman Riaz et al.}

\institute{
    Smart Data and Knowledge Services (SDS), German Research Center for Artificial Intelligence GmbH (DFKI), Trippstadter Straße 122
    67663 Kaiserslautern\\\email{\{firstname.lastname\}@dfki.de}\\ \and
    Department of Computer Science, RPTU Kaiserslautern-Landau, Erwin-Schrödinger-Straße 52, 67663 Kaiserslautern, Germany\and
    DeepReader GmbH, 67663 Kaiserlautern, Germany\\
}

\maketitle              %
\begin{abstract}

In this study, we introduce StylusAI, a novel architecture leveraging diffusion models in the domain of handwriting style generation. StylusAI is specifically designed to adapt and integrate the stylistic nuances of one language's handwriting into another, particularly focusing on blending English handwriting styles into the context of the German writing system. This approach enables the generation of German text in English handwriting styles and German handwriting styles into English, enriching machine-generated handwriting diversity while ensuring that the generated text remains legible across both languages. To support the development and evaluation of StylusAI, we present the \lq{Deutscher Handschriften-Datensatz}\rq~(DHSD), a comprehensive dataset encompassing 37 distinct handwriting styles within the German language. This dataset provides a fundamental resource for training and benchmarking in the realm of handwritten text generation. Our results demonstrate that StylusAI not only introduces a new method for style adaptation in handwritten text generation but also surpasses existing models in generating handwriting samples that improve both text quality and stylistic fidelity, evidenced by its performance on the IAM database and our newly proposed DHSD. Thus, StylusAI represents a significant advancement in the field of handwriting style generation, offering promising avenues for future research and applications in cross-linguistic style adaptation for languages with similar scripts.

\end{abstract}

\section{Introduction}
Despite significant technological advancements in our society, the use of traditional handwritten text remains widely popular for documenting data, making the task of handwritten text recognition (HTR) critically important for automated document processing.
However, extensive data complexity in handwritten texts, such as varying writing styles and languages~\cite{Grosicki:2011,Kleber:2013}, low-quality images~\cite{Aradillas:2021,Lins:2011,Pratikakis:2018}, and lighting variations~\cite{Lins:2011,Pratikakis:2018,Pratikakis:2012}, makes HTR a challenging task.
While recent Deep Learning (DL)-based systems have shown promising potential for improvement~\cite{li:2023,riaz:2022,maqsood:2023}, these models are mostly data-driven, especially transformer models, known for requiring extensive data for optimal performance. 
On the other hand, manually gathering and annotating handwritten text is an extremely labor-intensive and time-consuming task, requiring significant human effort~\cite{Grosicki:2011,Kleber:2013}. 

Due to the aforementioned challenges in gathering handwritten text data, augmenting these datasets through image synthesis is seen as a popular alternative~\cite{graves:2013,bhunia:2021}. Numerous synthesis and data augmentation approaches have been proposed in recent years~\cite{bhunia:2021,Luhman:2020,Nikolaidou:2023,Rombach:2022}, leading to significant improvements in the performance of existing HTR models~\cite{li:2023,pippi:2023}.
These include the previous state-of-the-art (SotA) approaches based on Generative Adversarial Network (GANs)~\cite{kang:2020,kang:2021,mattick:2021,fogel:2020} and the recent SotA approaches based on Diffusion Model (DMs)~\cite{Luhman:2020,Nikolaidou:2023,Zhu_2023_CVPR}, both of which typically synthesize handwritten text images by conditioning the generation process on the target text.
While the most recent DM-based synthesis methods~\cite{Nikolaidou:2023,Zhu_2023_CVPR} have significantly improved over the previous SotA approaches~\cite{kang:2020,kang:2021,mattick:2021,fogel:2020}, there is still a significant lack of research that explores style adaptation between similarly written languages. 
This research gap is particularly important, especially for enhancing the synthesis of handwritten text for languages with limited resources. Specifically, resource-constrained languages could benefit from adopting diverse styles from well-resourced languages that share similar scripts. In this paper, we aim to investigate this possibility for the German language by exploring style adaptation from English to German.

\begin{figure}[b!]
\begin{center}
  \includegraphics[width=1\columnwidth]{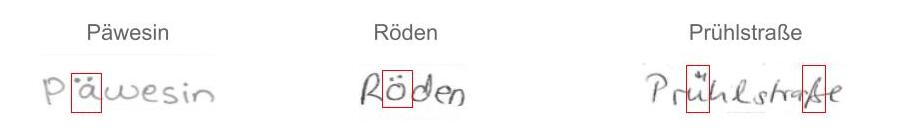}
  \caption[]{Few images showcasing a selection of written German words, with a particular focus on the unique characters found in the German alphabet, including umlauts and the eszett.}
  \label{fig:german_char}
\end{center}
\end{figure}

Despite the similarity between the German and English languages in their written forms, there exist specific German characters (as shown in Fig.~\ref{fig:german_char}) such as \textbf{\"a}, \textbf{\"o}, \textbf{\"u}, and \textbf{\ss}, that do not exist in the English vocabulary. 	
In addition, there is a huge scarcity of publicly available German handwritten text datasets, making it difficult to achieve sufficient diversity when synthesizing German handwritten text using existing data-driven text synthesis methods. 	
While one may train existing approaches on publicly available English language datasets, such as the IAM Handwriting Database~\cite{marti:2002}, to generate text for 
overlapping characters between the two languages (English and German), such an approach will naturally fail to generate text for out-of-vocabulary characters, such as those mentioned above. 
In this work, therefore, our main focus is to explore the possibility of adapting writing styles from large-scale English language datasets to the out-of-vocabulary German characters, so that small-scale German language datasets may then be augmented to produce diverse, style-rich datasets.  We achieve this by designing a conditional diffusion model, the generation process of which is guided not only through text and writer style but also by an additional synthetic printed text image. This addition of a visual representation of the target text allows us to model the problem as an image-to-image translation task, which helps improve style adaptation compared to existing text-only conditional approaches.

The main contributions of this paper are three-fold:

\begin{enumerate}
    \item We present the \lq{Deutscher Handschriften-Datensatz}\rq~(DHSD), a German handwriting dataset that comprises 37 distinct handwriting styles. 
    \item We propose StylusAI, an architecture based on diffusion models for handwritten text generation, for effective style adaptation incorporating the stylistic elements present in English handwriting into the context of the German writing system, creating a fusion that maintains legibility and coherence for both languages. This allows German to be generated in English writer styles and vice versa leading to diverse handwriting style generation. 
    \item Furthermore, we show that the proposed model also outperforms previously suggested models on datasets such as IAM and our new DHSD in terms of producing handwriting samples with superior text and stylistic quality.
\end{enumerate}

The rest of the paper is structured into the following main sections. In Section~\ref{sec:relatedwork}, a summary of related work in the field of handwritten text generation is presented. Section~\ref{sec:methodology} explores the proposed technique for the task at hand. Section~\ref{sec:experiments} details the experimental setup, including preprocessing steps and implementation details. Section~\ref{sec:results} presents the findings and their interpretation. Finally, Section~\ref{sec:conclusion} concludes the study and outlines future research directions.

\section{Related Work}
\label{sec:relatedwork}

Handwritten text generation holds immense significance in the field of document analysis and recognition and has been widely explored in the past decade \cite{graves:2013,kang:2020,mattick:2021,fogel:2020,Zhu_2023_CVPR}. The earliest attempts in this domain involved using Recurrent Neural Networks (RNNs) for online handwriting generation~\cite{graves:2013}. In particular, Graves~\cite{graves:2013} proposed using Long Short-Term Memory (LSTM) networks~\cite{Hochreiter:1997} to predict text-conditioned real-valued data sequences for synthesizing handwritten text in an online fashion.

More recently, the field has greatly shifted its attention towards offline text generation using generative approaches~\cite{kang:2020,mattick:2021,Zhu_2023_CVPR}.
Kang~\etal~\cite{kang:2020} proposed GANwriting for handwritten text generation, a conditional Generative Adversarial Network (GAN)~\cite{gans} that incorporated handwriting image samples of the writers for style information alongside the target text conditioning to guide the generation process.
While the approach achieved great success in generating realistic handwritten text images at the word level (later also extended for sentence-level generation~\cite{kang:2021}), allowing for a controlled generation of previously unseen text sequences with various writer styles, it greatly suffered from unrealistic pen-level artifacts introduced in the generated images. 
This issue was addressed in a follow-up work, SmartPatch~\cite{mattick:2021}, where an additional patch-level discriminator loss was employed to improve the generation quality. Fogel~\etal~\cite{fogel:2020} proposed ScrabbleGAN, where a GAN was trained in a semi-supervised fashion to generate comprehensive handwritten text sentences across various styles and content.
In a different direction, Bhunia~\etal~\cite{bhunia:2021} proposed an encoder-decoder transformer architecture that was trained to learn not only the long and short-range contextual relationships but also the style-content relationship through self-attention~\cite{vaswani:2017}.

The recent success of diffusion models in natural image synthesis~\cite{Ho:2020,Rombach:2022,imagen} has also sparked an interest in using them for the task of handwritten text generation. 
Luhman~\etal~\cite{Luhman:2020} recently proposed a conditional diffusion model for generating handwritten text sequences in an online fashion. 	
In particular, their approach uses diffusion to generate real-valued pen stroke sequences and introduces style and textual conditioning through embeddings generated by a pre-trained MobileNetV2 model~\cite{howard2017mobilenets}. It is important to highlight that while their approach utilized diffusion, it focused solely on online generation, which contrasts with the offline generation focus of this work. In another recent work, Nikolaidou~\etal~\cite{Nikolaidou:2023} proposed WordStylist, a UNet~\cite{ronneberger:2015}-based text-to-image latent diffusion model~\cite{Rombach:2022} for offline generation of handwritten text. 
In particular, they utilized a transformer-based architecture to generate character-level text embeddings and incorporated a cross-attention mechanism into the UNet model to condition the generation process on the target text and style embeddings.
It is worth mentioning that although WordStylist~\cite{Nikolaidou:2023} demonstrated exceptional performance in generating realistic handwritten text sequences, significantly outperforming the previous state-of-the-art~\cite{kang:2020,mattick:2021,bhunia:2021}, it was not trained to achieve cross-language style adaptation, which is the main focus of this work. Later, we demonstrate through results that WordStylist~\cite{Nikolaidou:2023} struggles to adapt styles from English to German when queried to generate out-of-vocabulary German characters in the style of English writers, despite being trained on a dataset that includes both English and German texts.

\section{Background}
\label{sec:methodology}
\subsection{Diffusion Models}
Diffusion models represent a class of generative models utilizing Markov chains to systematically introduce noise, thus obscuring the data's original structure. These models are tasked with learning how to invert this process, aiming to restore the data to its initial form. Their design is influenced by principles found in thermodynamics~\cite{dickstein:2015}, and these models have become increasingly prominent in image synthesis for their capability to produce high-fidelity images. The Diffusion Model consists of two main phases: the forward process, which involves the diffusion of a sample, and the reverse process, which involves denoising~\cite{Ho:2020}.

\subsubsection{Forward Diffusion}
In the forward diffusion process, an initial sample, denoted by $x_0$ is obtained from a distribution $q(x_0)$. This sample is subsequently perturbed with Gaussian noise to produce a latent variable $x_1$. The procedure of introducing noise and creating subsequent latent variables ($x_2$, $x_3$, \ldots, $x_T$) is repeated until it reaches a predetermined hyperparameter T.
Mathematically, the relationship between the latent variables is expressed as follows:
\begin{equation}
q(x_{1:T} | x_0) = \prod_{t=1}^T q(x_t | x_{t-1}), \quad q(x_t | x_{t-1}) \sim \mathcal{N}(x_t; \sqrt{1 - \beta_t}x_{t-1}, \beta_t I)
\label{eq:diffusion}
\end{equation}
where we have $\beta_i \in [0, 1]$ for all $i \in [1, T]$. $\beta_{1}, \beta_{2}, ..., \beta_{T}$ form a noise variance schedule and dictate the quantity of noise added at every timestep. In the concluding step, assuming a sufficiently large \(T\) and an appropriate noise schedule, we will have \(x_T\) that closely resembles a sample drawn from pure Gaussian noise.

\subsubsection{Reverse Diffusion}
During the denoising phase, a U-Net architecture is employed to iteratively mitigate noise emanating from a normal distribution, with the objective of restoring the original dataset. The process of image generation relies on a sequential sampling technique. It starts by generating a sample from \(q(x_T)\). Then, a sample is taken from the distribution at the previous timestep, conditional on the value of \(x_T\), and this procedure is repeated in reverse order until reaching \(x_0\).

The noise reduction across reverse timesteps is guided by the transition:
\begin{equation}
p_\theta(x_{0:T}) = p(x_T) \prod_{t=1}^T p_\theta(x_{t-1}|x_t), \hspace{5mm} p_\theta(x_{t-1}|x_t) = \mathcal{N}(x_{t-1}; \mu_\theta(x_t, t), \Sigma_\theta(x_t, t)).
\end{equation}
The network is trained by minimizing the variational lower bound between the posterior of the forward process and the joint distribution of the reverse process, represented as \( p_\theta \).

The training loss is defined as:

\begin{equation}
\label{eq:loss}
\mathcal{L} = \mathbb{E}_{\mathbf{x}, t, \epsilon \sim \mathcal{N}(0,1)} \left[|| \epsilon - \epsilon_\theta(\mathbf{x}_t, t) ||^2_2 \right]
\end{equation}

where this loss quantifies the discrepancy between the true noise \( \epsilon \) and the noise predicted by the network \( \epsilon_\theta \).

\section{StylusAI: The Proposed Approach}
 Inspired by the work of Brooks~\etal~\cite{brooks:2023}, instead of relying solely on simple text conditioning as done in previous work~\cite{Nikolaidou:2023}, we enhance our method by incorporating synthetic printed text image conditioning during the diffusion process. We approach the problem of style adaptation by modeling a diffusion process whose goal is to generate a handwritten text image from a printed text image, adhering to a specific style and text. The printed text image provides details on the appearance of the characters, and the diffusion process meticulously generates those characters in the targeted writer's style. This innovative approach allows for better style adaptation of German characters in English writer styles and vice versa. Given a handwriting image \(x\), the diffusion process aims to add noise to it, producing a noisy image \(x_t\) where the noise level increases over timesteps \(t \in T\). We propose a U-Net based network \(\epsilon_{\theta}\) that predicts the noise added to \(x_t\) given the text conditioning \(c_T\), writer style conditioning \(c_Y\), and printed text image conditioning \(c_P\). The following diffusion objective is minimized to train the network:

\begin{equation}
\label{eq:loss}
\mathcal{L} = \mathbb{E}_{\mathbf{x}, \mathbf{c}_T, \mathbf{c}_Y, \mathbf{c}_P, \epsilon \sim \mathcal{N}(0,1)} \left[|| \epsilon - \epsilon_\theta(\mathbf{x}_t, t, \mathbf{c}_T, \mathbf{c}_Y, \mathbf{c}_P) ||^2_2 \right]
\end{equation}

\begin{figure}[t!]
\centering
\includegraphics[width=\textwidth]{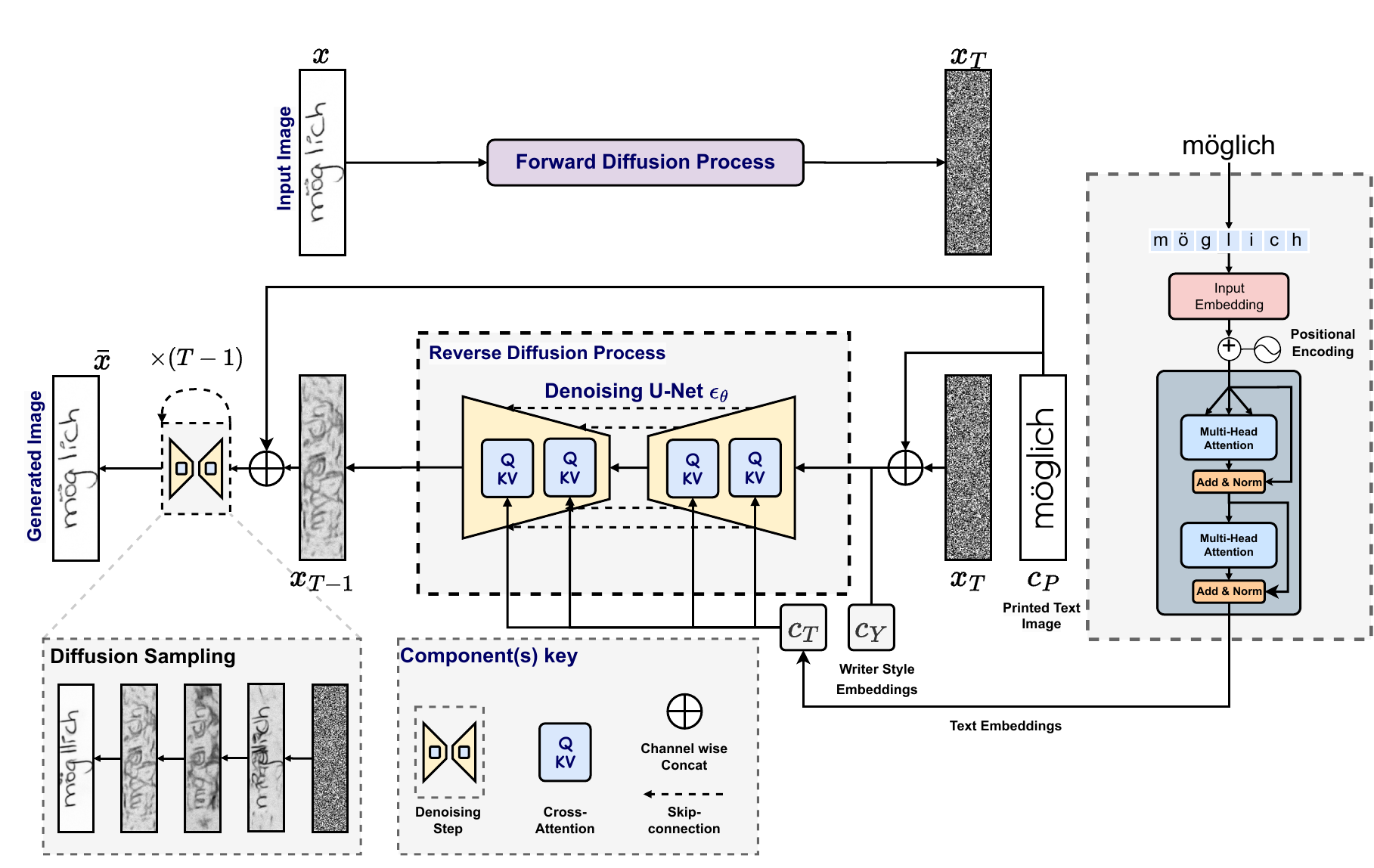}
\caption{Overview of the Proposed Architecture.} 
\label{fig:model}
\end{figure}

StylusAI leverages synthetically generated printed text images, character-level text, and writer-style embeddings to facilitate controlled handwritten text generation. It employs a standard Transformer Encoder~\cite{vaswani:2017} for text conditioning. For predicting noise, StylusAI utilizes a U-Net~\cite{ronneberger:2015} based model as illustrated in the Figure~\ref{fig:model}. The addition of printed text image in the conditioning allows for better style adaptation between similarly written languages as evidenced from results described in Section~\ref{sec:results}. The details of the architecture are given below.

\subsection{Transformer Encoder and Style Conditioning}
We use character-level tokenization for text conditioning and pass the tokens through two stacks of transformer encoder layers, which utilize multi-head self-attention layers to generate text embeddings. The attention mechanism is defined as $\text{Att}(Q, K, V) = \text{softmax}\left(\frac{QK^T}{\sqrt{d_k}}\right)V$ which gives a weighted sum of character embedding representation. These embeddings are used for conditioning via cross-attention with features from U-Net layers. The writer style condition \(c_Y\) is processed through an embedding layer and subsequently added to the time step embedding as shown in Figure~\ref{fig:model}.

\subsection{Forward Diffusion and Training}
We aim to train a diffusion model \(p_{\theta}(x \,|\, c_T, c_Y, c_P)\), given the input text \(c_T\), writer style \(c_Y\), and a synthetic printed text image \(c_P\). The content of the input text \(c_T\), which is to be conditioned on, matches that of the printed text image. We sample the timesteps \(t \in T\) from a uniform distribution which are encoded using a sinusoidal position embedding. Noise is incrementally added to the original image \(x\). A noise scheduler incrementally increases the level of noise from \(\beta_1 = 10^{-4}\) to \(\beta_T = 0.02\) over \(T = 1000\) timesteps. To predict the noise \(\epsilon\), we utilize a U-Net architecture that comprises Residual Blocks~\cite{he:2016} and intermediate cross-attention Transformer blocks~\cite{Rombach:2022}. The cross-attention blocks are employed to attend to the text conditioning. The network accepts as input the noisy image, the corresponding timestep, and the specified conditions \(c_Y\), \(c_T\), and \(c_P\) (i.e., the printed text image concatenated channel-wise with the noisy image). The training goal is to minimize the discrepancy between the noise prediction of the network and the actual noise in the image mathematically described in Equation~\ref{eq:loss}.

\subsection{Reverse Diffusion: Denoising}
For the generation process, we employ the learned diffusion model for reverse diffusion. We feed a noisy random sample \(x_T\) into the learned network \(\epsilon_{\theta}\) along with a channel-wise concatenation of the printed text image \(c_P\), text condition \(c_T\), and style condition \(c_Y\) to predict the noise. The predicted noise is progressively removed at each timestep, repeating this process from \(T\) to \(t = 0\), until we obtain our handwritten text image sample. We use the DDPM algorithm for the denoising process. The denoising process is graphically shown in Figure~\ref{fig:model}.

\section{Experimental Setup}
\label{sec:experiments}

\subsection{Datasets}
\label{sec:datasets-used}
We employ two different datasets, the IAM dataset~\cite{marti:2002} and the proposed DHSD dataset. The specifics of the mentioned dataset including the splits used for training are given below:

\subsubsection{IAM:}
\begin{figure}[t]
\begin{center}
  \includegraphics[width=1\columnwidth]{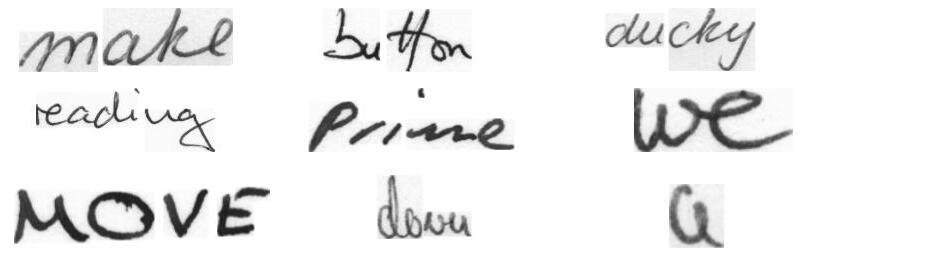}
  \caption[]{These images provide a glimpse into the diversity of IAM dataset, showcasing a variety of unique handwriting styles}
  \label{image:german1}
\end{center}
\end{figure}

The IAM Handwriting Database is a comprehensive resource designed for use in handwriting recognition and related research. It encompasses a collection of handwriting samples provided by 657 different writers, ensuring a diverse representation of handwriting styles. In total, the database contains 1,539 pages of scanned text, which further breaks down into more granular elements including 5,685 isolated and labeled sentences. Researchers can also access 13,353 isolated and labeled text lines for studying specific handwriting characteristics at the line level. At the most detailed level, the database features a substantial compilation of 115,320 isolated and labeled words, offering an in-depth opportunity for word recognition analysis. These words were extracted from the scanned pages using an automatic segmentation method, which was subsequently followed by a thorough manual verification process to ensure the accuracy and reliability of the segmentation and labeling. We use labeled word images in our experiments and employ the Aachen splits, similar to~\cite{Nikolaidou:2023}, for training purposes. Figure~\ref{image:german1} displays sample images of handwritten words from the IAM database.

\subsubsection{Deutscher Handschriften-Datensatz (DHSD):}
\begin{figure}[ht]
\begin{center}
  \includegraphics[width=1\columnwidth]{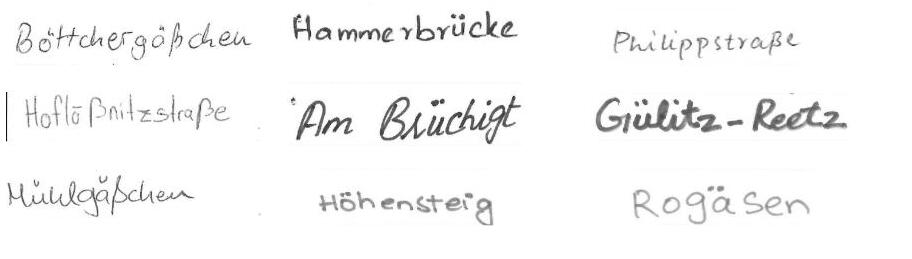}
  \caption[]{These images provide a glimpse into the diversity of our proposed DHSD, showcasing a variety of unique handwriting styles.}
  \label{image:german}
\end{center}
\end{figure}

We propose a novel German handwriting dataset, which has been populated with contributions from 37 individuals, each providing an average of 150 words. To guarantee the representation of the entire German alphabet within the dataset, we carefully selected words to ensure that each one includes at least one letter from the German alphabet. The dataset comprises words that are names of cities and streets in Germany, all extracted from the OpenStreetMap (OSM) database.
The dataset is split into training and testing subsets, with 80\% allocated for training and 20\% for testing. Sample images of the handwritten words from our proposed dataset are displayed in Figure~\ref{image:german}.

\subsubsection{IAM + DHSD}
For the purpose of evaluating our model's ability to adapt German characters in different English writing styles, we integrated the IAM dataset with our newly proposed DHSD. Specifically, we compiled data from 37 English writers within the IAM dataset and paired it with data from 37 German writers from our proposed dataset. This merged dataset was employed for training our model. The top 37 writers who contributed the most writing samples were selected from the IAM dataset with an average of 326 words per writer. The experimentation on this combined dataset is conducted differently compared to the datasets mentioned above. The outcomes of this integration are elaborated upon in Section~\ref{sec:results}.

\subsection{Experimental  Setup}
Our experimental setup is largely based on the one described by~\cite{Nikolaidou:2023}, with modifications to accommodate our newly proposed DHSD dataset. We evaluate the model's performance by assessing the accuracy of both handwriting text recognition and writer style recognition on the handwritten text images produced by the generative model. We intentionally avoid using the Fréchet Inception Distance (FID) metric. Although FID is a common metric for assessing generative models, it might not be suitable for tasks that deal with image types substantially different from the natural images seen in ImageNet, which was used to train the underlying FID network. This discrepancy between the types of images can undermine the reliability of the evaluation process. However, modifying the metric by fine-tuning the FID network on a dataset of document images goes beyond the scope of this paper.
\subsection{Implementation Details}
Our proposed architecture is implemented using a U-Net model, which includes five encoder and decoder blocks. The initial 2 blocks and the last block in the encoder are ResNet layers with downsampling. We use one Resnet layer per each U-Net block. The remaining two blocks feature cross-attention layers with 4 heads for Multi-Head attention. Conversely, the U-Net's decoder segments mirror this configuration by employing upsampling, following the same sequence in reverse. During the denoising process, we append a synthetically generated printed text image to the noisy image at each timestep. This printed text image, which corresponds to the text designated for generation, is appended along the channel dimension. The timestep and writer-style embeddings are both kept at a dimension of 256. The proposed architecture was trained using four A100-40GB GPUs, with a collective batch size of 256. All text images were resized to 256 x 64 pixels in grayscale, maintaining the aspect ratio. We trained and evaluated the architecture across three different dataset configurations as detailed in Subsection~\ref{sec:datasets-used}.

\section{Results and Analysis}
\label{sec:results}

We conduct our assessment following the evaluation framework outlined in~\cite{Nikolaidou:2023}, spanning two dimensions: textual quality and style quality for IAM and DHSD. For textual quality, we create synthetic training sets (generated by the models) like the training splits of the dataset configurations IAM and DHSD as detailed in Section~\ref{sec:experiments}. We use these sets to train a handwriting recognition model~\cite{riaz:2022}, which is then evaluated on the original test sets to determine the model's performance on individual datasets. Meanwhile, style quality is gauged by training a conventional CNN on IAM and DHSD to classify writers' styles. The effectiveness of this approach is tested on the generated handwriting samples.

For the assesment of style adaptation we also use textual quality and style quality but the experimentation is done a bit differently. The evaluation process differs in this case, as we utilize a combined dataset comprising both the IAM and DHSD datasets for training the generative model. Subsequently, the trained model is employed to generate a modified version of the combined dataset, where the styles are reversed. This entails generating German words in the style of 37 English writers with a one-to-one correspondence between words and writers.

\subsection{Handwriting Text Recognition (HTR)}
We evaluate text quality using a handwriting text recognition model (HTR) proposed by~\cite{riaz:2022}, which we train on synthetically generated training sets of IAM and DHSD. This evaluation approach is similar to what is proposed by the authors of WordStylist~\cite{Nikolaidou:2023}. These sets were synthesized utilizing previous methods such as GANwriting,  Smart Patch, Word Stylist, and our proposed approach for comparison. The Handwritten Text Recognition (HTR) model, once trained, undergoes evaluation on the original test splits from both datasets using the Character Error Rate (CER) as a measure of text quality. A lower CER signifies that the HTR model has effectively generalized to the test set, suggesting that the synthetically generated words used in training are legible and closely resemble the original test set distribution. This similarity is the intended result of a good handwritten text generation model. The evaluation results for text quality on the IAM dataset are presented in Table~\ref{table:iam}, and on the DHSD dataset in Table~\ref{table:dhsd}. Our proposed model outperforms previous existing models in both cases.

\begin{table}[t]
  \centering
  \renewcommand{\arraystretch}{1.2} %
  \setlength{\tabcolsep}{10pt} %
  \caption{The HTR results for the IAM dataset using the Character Error Rate (CER), where a lower rate signifies improved performance.}
  \label{table:iam}
  \begin{tabular}{l r} %
    \toprule
    \textbf{Training Data} & \multicolumn{1}{r}{\textbf{CER(\%)}} \\ %
    \midrule
    Real IAM & 4.57 ± 0.07 \\
    GANwriting IAM & 35.21 ± 0.23 \\
    SmartPatch IAM & 30.25 ± 0.45 \\
    WordStylist IAM & 8.50 ± 0.12 \\
    StylusAI IAM (Ours) & 7.82 ± 0.09 \\
    Real IAM + WordStylist & 4.42 ± 0.08 \\
    Real IAM + StylusAI IAM (Ours) & \textbf{3.85 ± 0.09} \\
    \bottomrule
  \end{tabular}
\end{table}

\begin{table}[t]
  \centering
  \renewcommand{\arraystretch}{1.2} %
  \setlength{\tabcolsep}{10pt} %
  \caption{The HTR results for the DHSD dataset using the Character Error Rate (CER), where a lower rate signifies improved performance.}
  \label{table:dhsd}
  \begin{tabular}{l r} %
    \toprule
    \textbf{Training Data} & \multicolumn{1}{r}{\textbf{CER(\%)}} \\ %
    \midrule
    Real DHSD & 11.13 ± 0.07 \\ %
    GANwriting DHSD & 42.31 ± 0.13 \\
    SmartPatch DHSD & 37.47 ± 0.25 \\
    WordStylist DHSD & 14.58 ± 0.12 \\
    StylusAI DHSD (Ours) & 11.57 ± 0.09 \\ %
    Real DHSD + WordStylist DHSD & 11.62 ± 0.08 \\ 
    Real DHSD + StylusAI DHSD (Ours) & \textbf{9.01 ± 0.08} \\ %
    \bottomrule
  \end{tabular}
\end{table}

The authors of WordStylist propose this method of evaluating textual quality, which we have chosen to use as-is for IAM and DHSD datasets. However, we recognize that it may not accurately reflect the generative capabilities of a generative model. Generating the same training set on which the model was originally trained could result in good-quality images, but this may be due to overfitting. To effectively assess the extent of style adaptation of the generative model we follow a different approach.

\subsubsection{HTR for style adaptation}
To assess the effectiveness of style adaptation in text quality, we employ the combined IAM+DHSD dataset to train the generative model. Once the model is trained, we use it to generate the DHSD training split in the styles of 37 English writers found in the combined dataset. This newly created synthetic training data (Eng-DHSD) is then used to train the HTR model and it does not depict the original training distribution as German words are not present in the style of English writers during the training of the generative model. The performance of the HTR model is evaluated on the test split of the DHSD dataset. A lower CER would indicate a more successful style adaptation. Only WordStylist and StylusAI are considered for comparison because GANwriting and SmartPatch did not yield promising results for this task. The comparison suggests that our model outperforms WordStylist as shown in Table~\ref{table:eng-dhsd}.

\begin{table}[t]
  \centering
  \renewcommand{\arraystretch}{1.2} %
  \setlength{\tabcolsep}{10pt} %
  \caption{The HTR results after training on Eng-DHSD dataset and testing on DHSD testing split using the Character Error Rate(CER), where a lower rate signifies improved performance.}
  \label{table:eng-dhsd}
  \begin{tabular}{l r} %
    \toprule
    \textbf{Training Data} & \multicolumn{1}{r}{\textbf{CER(\%)}} \\ %
    \midrule
    Real DHSD & 11.13 ± 0.07 \\
    WordStylist Eng-DHSD & 45.80 ± 0.12 \\
    StylusAI Eng-DHSD (ours)& 30.86 ± 0.07 \\
    Real DHSD + WordStylist Eng-DHSD & 12.67 ± 0.08 \\
    Real DHSD + StylusAI Eng-DHSD (Ours) & \textbf{10.24 ± 0.09} \\
    \bottomrule
  \end{tabular}
\end{table}

\begin{figure}[b!]
\begin{center}
  \includegraphics[width=1\textwidth]{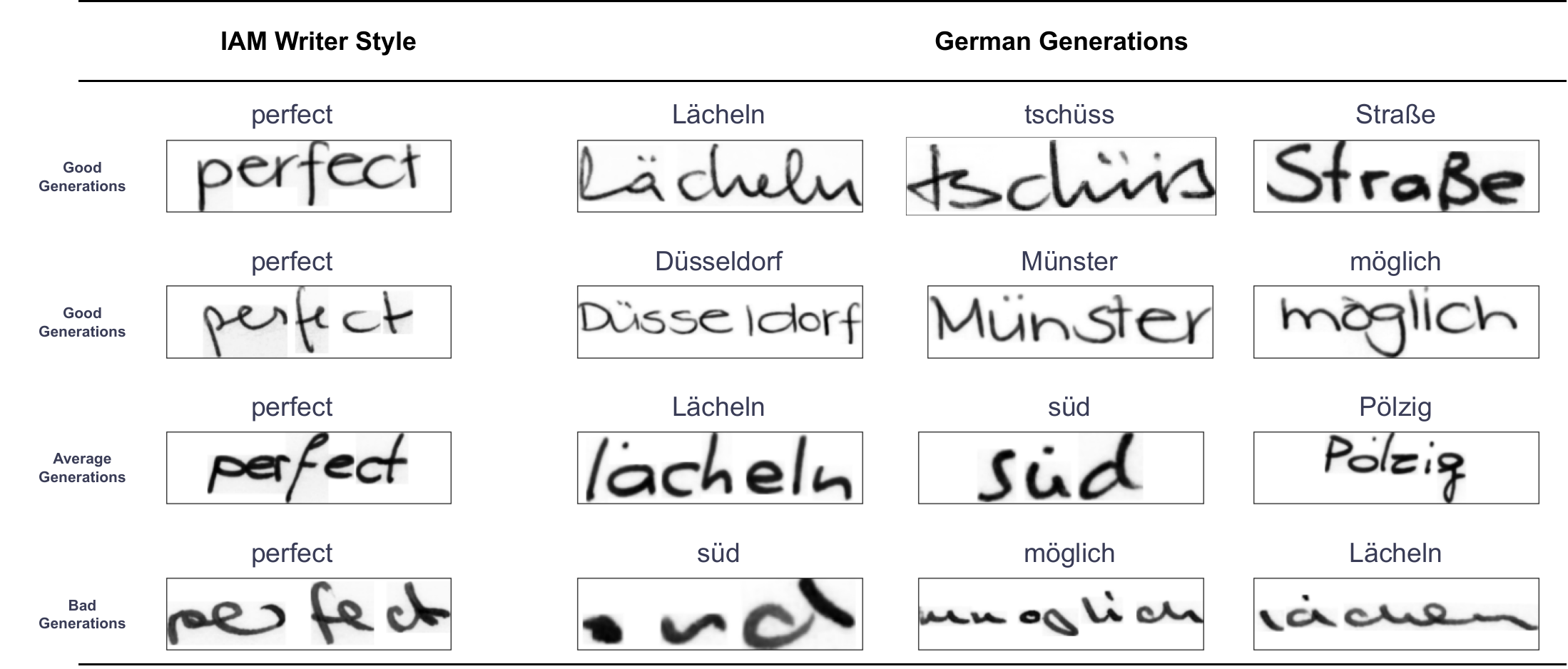}
  \caption[]{German text samples generated by StylusAI, emulating different IAM writer styles, where good generations are those in which the German characters have been better adapted to the English writer styles compared to average and poor generations.}
  \label{image:error_results}
\end{center}
\end{figure}

Generating German characters in the style of English writers is a difficult task for a generative model that has been trained on handwritten text images without examples of German characters written in the style of English writers. Consequently, when it comes to generating German characters in the style of English writers, the model produces inconsistent results. The consistency of these generations varies among different English writer styles. Please refer to Figure~\ref{image:error_results} for visual examples.

\begin{figure}[ht]
\begin{center}
  \includegraphics[width=1\textwidth]{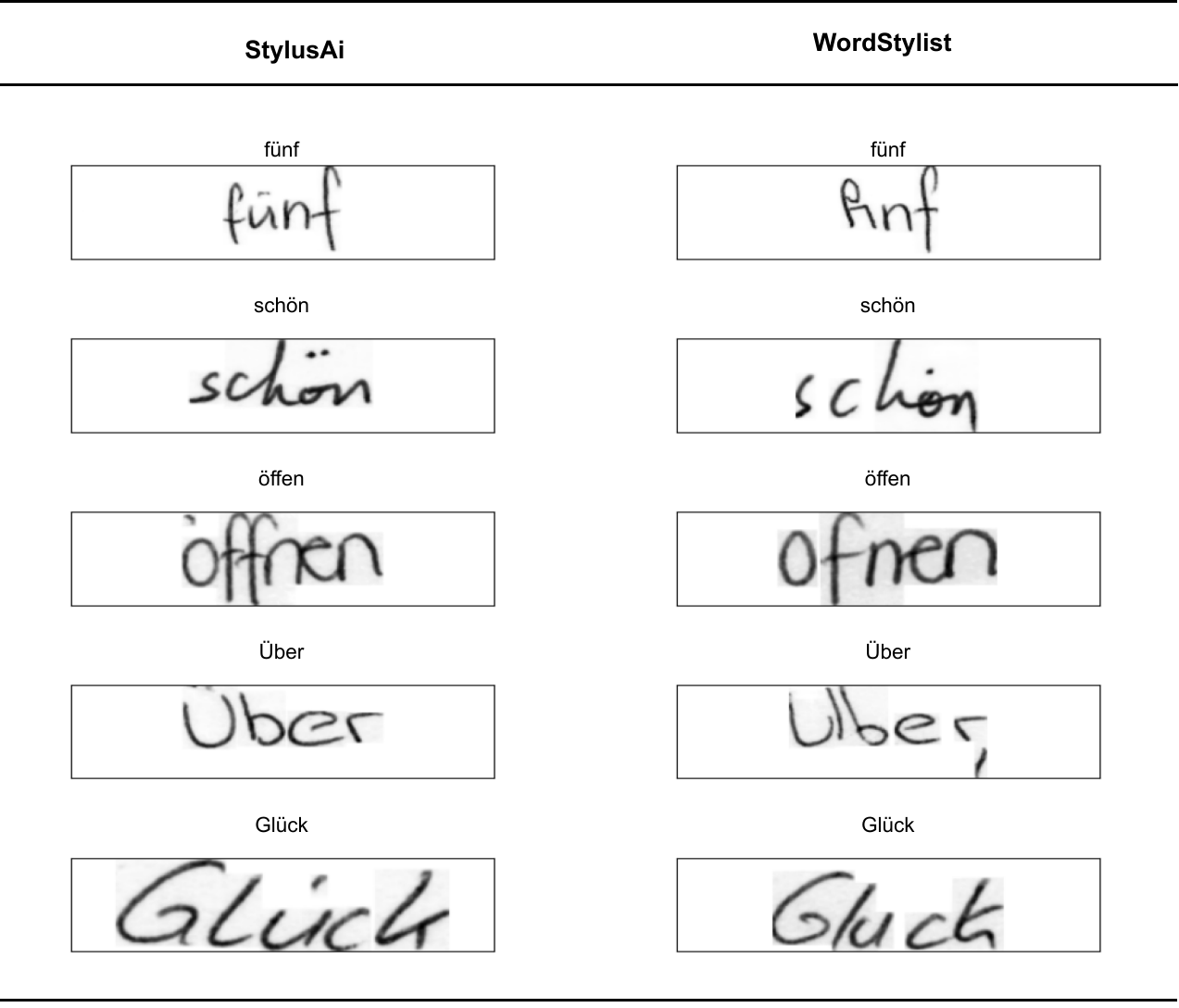}
  \caption[]{Comparison between German word generations from StylusAi and WordStylist in IAM English writer styles. StylusAI demonstrates superior quality in adapting German characters to English writer styles.}
  \label{image:comparison}
\end{center}
\end{figure}

Despite the inconsistencies, StylusAI consistently produces better results compared to WordStylist across various writing styles. The incorporation of a printed text image, along with character embeddings, during the denoising process provides additional guidance on the basic style of individual characters, resulting in more consistent outcomes. Figure~\ref{image:comparison} provides a comparison between generations of WordStylist and StylusAI (ours). 

\subsection{Writing Style Analysis}
To assess the model's effectiveness in capturing the diverse writing styles of different writers, we followed an evaluation methodology similar to the one outlined in~\cite{Nikolaidou:2023}. To evaluate the generated styles, we fine-tuned a ResNet-18 CNN~\cite{he:2016}, originally pre-trained on ImageNet, utilizing the IAM and DHSD databases for writer classification tasks. Following this, we utilized the datasets produced by the four generative approaches as test sets and presented the resulting accuracy. For analysis of writer style classification on style adaptation task, we utilize the same pattern by training on IAM+DHSD and testing on the Eng-DHSD synthesized set from the trained generative models. The evaluation for writer classification is shown in Table~\ref{table:combined_accuracy}. The results show that StylusAI is able to adapt the style better while also producing fewer errors while generating German characters.

\begin{table}[ht]
  \centering
  \renewcommand{\arraystretch}{1.2} %
  \setlength{\tabcolsep}{5pt} %
  \caption{Comparison of the accuracy of a ResNet18 model trained for writer identification on IAM, DHSD, and IAM+DHSH datasets and tested on the generated IAM, DHSD and Eng-DHSD datasets.}
  \label{table:combined_accuracy}
  \begin{tabular}{l r r r} %
    \toprule
    \textbf{Method} & \textbf{IAM (\%)} & \textbf{DHSD (\%)} & \textbf{Eng-DHSD (\%)} \\ 
    \midrule
    GANwriting & 4.81 & 6.72 & - \\
    SmartPatch & 4.09 & 7.28 & - \\
    WordStylist & 70.67 & 73.50 & 62.38 \\
    StylusAI (Ours) & \textbf{75.25} & \textbf{75.02} & \textbf{66.79} \\
    \bottomrule
  \end{tabular}
\end{table}

\section{Conclusion and Future Work}
\label{sec:conclusion}
This research introduces the \lq{Deutscher Handschriften-Datensatz}\rq (DHSD), a comprehensive dataset encompassing a wide array of German handwriting styles, laying the groundwork for novel applications in German handwriting analysis and generation. Leveraging this dataset, we developed StylusAI, a state-of-the-art architecture premised on diffusion models, tailored for the intricate task of handwriting style adaptation. StylusAI represents a significant stride forward, that combines stylistic elements prevalent in English handwriting with those inherent to the German writing system. This amalgamation not only preserves but enhances the legibility and stylistic cohesion across both languages, promoting a seamless generation of diverse handwriting styles. Our extensive evaluations demonstrate that StylusAI not only achieves but surpasses the performance benchmarks of existing models in the realm of handwritten text generation. Its capabilities are evident when assessed on both the newly curated DHSD and the established IAM datasets, where it consistently generates handwriting samples of superior text and stylistic quality. 

This paper signifies the immense potential of employing diffusion models in the context of cross-linguistic handwriting synthesis between similarly written languages, paving the way for advancements in the field of handwritten text generation. One interesting future direction includes the exploration of other similarly written languages to enhance handwritten text generation and consequently handwriting text recognition systems.

\bibliographystyle{splncs04}

\end{document}